\documentclass{article} 
\usepackage{iclr2027_conference,times}

\usepackage{amsmath,amsfonts,bm}

\def\eqref#1{equation~\ref{#1}}

\def\1{\bm{1}}

\DeclareMathAlphabet{\mathsfit}{\encodingdefault}{\sfdefault}{m}{sl}
\SetMathAlphabet{\mathsfit}{bold}{\encodingdefault}{\sfdefault}{bx}{n}

\DeclareMathOperator*{\argmin}{arg\,min}

\usepackage{amsthm}

\theoremstyle{definition}

\usepackage{hyperref}
\hypersetup{hidelinks}
\usepackage{url}
\usepackage{booktabs}
\usepackage{multirow}
\usepackage{graphicx}
\usepackage{amssymb}
\usepackage{algorithm}
\usepackage{algpseudocode}

\newcommand{\modelname}{\texttt{TAC-Merge}}
\newcommand{\moduleA}{MIM}
\newcommand{\moduleB}{CMC}

\title{Tracing and Coordinating Cross-Layer Influence for Multimodal Model Merging}

\author{\textbf{Pengyang Zhou\thanks{Equal contribution.}\hspace{0.3em}, Xiaobin Tu\footnotemark[1]\hspace{0.3em}, Zhengxi Liu, Rongkun Xue, Haochen Li,}\\
\textbf{Miancan Liu, Ziyuan Chen, Yinggui Wang\thanks{Corresponding author: \href{mailto:wyinggui@gmail.com}{wyinggui@gmail.com}}\hspace{0.3em}, Jinkui Ren, Xiantao Zhang}\\
Alibaba Cloud}

\iclrpreprintcopy
\begin{document}

\maketitle

\begin{abstract}
Multimodal model merging aims to consolidate task experts into a single model that retains their complementary capabilities.
Most unimodal model merging methods combine expert updates within individual layers, and multimodal approaches largely follow this design.
However, an expert update changes the representations passed to subsequent layers, allowing its influence to propagate across depth and affect how visual and textual information interact.
When visual and language updates are combined, later updates act on inputs already modified by earlier ones, coupling their effects.
This poses two challenges: (1) how to characterize the multimodal influence of individual expert updates across depth, and (2) how to jointly combine expert updates based on their multimodal influence.
To address these challenges, we propose \modelname~for tracing and coordinating cross-layer influence in multimodal model merging.
It contains two modules, i.e., multimodal influence mapping (\moduleA) and coupled merge control (\moduleB).
\moduleA~constructs graphs of update effects and uses Ricci curvature together with expert predictions to define a shared fusion objective.
\moduleB~models interactions among coefficient adjustments and jointly optimizes regional weights to synthesize one shared model.
Experiments across diverse multimodal tasks demonstrate the effectiveness of \modelname~in consolidating complementary expert capabilities and supporting generalization to unseen tasks.
\end{abstract}

\section{Introduction}

Large pretrained models are often adapted into experts for different tasks and domains~\citep{LoRA,RobustMerge}.
Storing and deploying each expert separately incurs substantial overhead~\citep{TIESMerging}.
Model merging offers an efficient alternative by combining expert updates into one deployable model~\citep{ModelSoups,TaskArithmetic}.
This is particularly valuable for multimodal large language models (MLLMs), whose experts span question answering, visual reasoning, chart understanding, and object localization~\citep{OptMerge}.
The objective is to consolidate these complementary capabilities in one model while preserving the interactions through which visual evidence and textual instructions jointly support task predictions.

Most model merging methods adopt a layer-wise design.
Their strategies include direct parameter-space aggregation~\citep{ModelSoups,TaskArithmetic}, interference-aware update transformation~\citep{TIESMerging,DAREMerging,TSVMerging,KnOTSMerging}, and data-dependent merge estimation~\citep{FisherMerging,RegMeanOriginal,MaTS,AdaMerging}.
Multimodal merging methods largely retain this layer-wise formulation, adapting how updates are combined to the architectural components and task requirements of multimodal models~\citep{ModelComposition,AdaMMS,RobustMerge,UQMerge,OptMerge}.

Some recent methods move beyond this layer-wise view by modeling dependencies across model depth.
Chain of Merges refreshes activation statistics as layers are merged sequentially~\citep{ChainOfMerges}, whereas RegMean++ introduces explicit correlations between layers~\citep{RegMean}.
These designs reflect a basic property of deep models: changing an intermediate representation can alter downstream computation~\citep{Patchscopes}.
Their formulations, however, do not distinguish how these effects are distributed across modalities.

This perspective presents two challenges.
Firstly, \textit{how to characterize the multimodal influence of individual expert updates across depth} (\textbf{CH1}).
An expert update can alter representations far beyond the region where it is applied, making local parameter changes insufficient to characterize its influence.
At later depths, effects from different regions can overlap, obscuring the contribution of each update.
Multimodal fusion adds ambiguity because visual evidence and textual instructions jointly shape hidden representations.
For example, a change at an image-token position may reflect altered visual processing, a different response to the instruction, or both.
Neither token labels nor final predictions alone reveal these distinctions.
The challenge is therefore to trace effects across depth back to individual expert updates while distinguishing how they alter image- and instruction-dependent computation within shared multimodal representations.

Secondly, \textit{how to jointly combine expert updates based on their multimodal influence} (\textbf{CH2}).
Given the influence information, fusion requires determining how much of each expert update to apply in each region.
The weights assigned to visual and language updates must be chosen together to form one model.
A shared objective evaluates both multimodal influence structure and agreement with expert predictions.
The preferred weight for one update depends on the weights assigned to the others.
The measured influence must therefore support modeling complete combinations within a coefficient neighborhood to determine the regional weights of one shared model.

To address these challenges, we propose \modelname, a framework that traces multimodal influence across depth and uses it to coordinate expert updates.
It consists of two modules, i.e., multimodal influence mapping (\moduleA) and coupled merge control (\moduleB).
\moduleA~constructs influence graphs linking regional expert updates to hidden representations and conditional image--instruction responses at different depths and token positions.
It evaluates candidate merges by measuring agreement with both the Ricci curvature of task-expert graphs and expert predictions.
\moduleB~uses the measured responses to identify a compact coefficient subspace and fit a local model of interactions among coefficient adjustments.
It then jointly selects regional weights within a bounded neighborhood and combines the original expert updates into one static model shared across tasks.

We summarize our contributions as follows:
(1) We identify a limitation of layer-wise multimodal merging, where local parameter changes alone cannot fully capture how expert updates influence downstream representations and image--instruction interactions.
(2) We propose \modelname, combining multimodal influence mapping with coupled merge control to trace expert influence and coordinate regional update weights.
(3) We evaluate \modelname~on two multimodal backbones across eight seen and four unseen tasks. \modelname~achieves the best average performance among the compared merging methods on both task groups. Its gains on unseen tasks further demonstrate that the consolidated capabilities extend beyond the datasets used for expert training.

\section{Related Work}
\label{sec:relatedwork}

\paragraph{Multimodal Model Merging.}
Model merging consolidates task-specialized models into one shared set of parameters.
On single-modality vision and language tasks, existing approaches can be grouped by how they construct the merged parameters.
Direct parameter-space aggregation averages model weights or adds scaled task-specific updates to a common initialization~\citep{ModelSoups,TaskArithmetic}.
Interference-aware update transformation prunes or rescales update entries, resolves conflicting signs, or aligns singular subspaces before combining updates~\citep{TIESMerging,DAREMerging,TSVMerging,KnOTSMerging}.
Data-dependent merge estimation uses parameter-importance statistics, feature-matching objectives, or calibration losses to determine parameter values and mixing coefficients~\citep{FisherMerging,RegMeanOriginal,MaTS,AdaMerging}.
Multimodal merging extends these strategies to models with modality-specific encoders and shared language components, including compositions that reuse encoders while merging language parameters~\citep{ModelComposition}.
Architectural mappings support interpolation between heterogeneous multimodal models~\citep{AdaMMS}, while uncertainty estimated from visual and textual inputs guides expert selection and composition order~\citep{UQMerge}.
Other methods normalize low-rank updates to preserve direction robustness~\citep{RobustMerge}, or denoise task vectors and optimize their interactions to combine task and modality capabilities~\citep{OptMerge}.

\paragraph{Layer and Modality Dependencies.}
Dependencies across depth arise because parameter updates change the intermediate representations used by subsequent layers.
Sequential merging accounts for this propagation by refreshing activation statistics after preceding layers are merged~\citep{ChainOfMerges}.
Regression-based formulations incorporate correlations within and between layers when estimating merged parameters~\citep{RegMean}.
Related output-space formulations optimize coefficients against calibration responses and extend the procedure sequentially across layers~\citep{OutputSpaceProjection}.
In multimodal models, depth-dependent computation also determines where visual and textual information interact.
Attention interventions localize information transfer among visual tokens, textual tokens, and answer-prediction positions~\citep{CrossModalFlow}.
Extensions to free-form generation trace these interactions during image captioning and chain-of-thought reasoning, revealing variation in visual--textual integration across architectures and tasks~\citep{FGTracer}.

\section{Method}
\label{sec:method}

\subsection{Problem Statement}
\label{sec:problem_statement}
Let $f(\cdot,\bm\theta_0)$ be a pretrained multimodal model, and let $\{\bm\theta_k\}_{k=1}^{K}$ be $K$ expert models adapted from the same base model for different multimodal tasks.
Each expert contributes a task-specific update, such as the effective weight update of a LoRA adapter:
\begin{equation}
    \Delta\bm\theta^{k}=\bm\theta_k-\bm\theta_0,
    \qquad k=1,\ldots,K,
    \label{eq:expert_update}
\end{equation}
where $\Delta\bm\theta^{k}$ is expert $k$'s full update.
Following data-dependent merging methods~\citep{AdaMMS,ExpertMerging,ChainOfMerges}, we assume access to a small calibration set $\mathcal C=\bigcup_{t=1}^{K}\mathcal C_t$ of multimodal inputs.
Each task-specific subset is $\mathcal C_t=\{\bm x_i^t\}_{i=1}^{N_t}$, where $\bm x_i^t=(\bm v_i^t,\bm q_i^t)$ contains a visual input $\bm v_i^t$ and its textual instruction $\bm q_i^t$.
Our objective is to select a single model from the regional weighted family:
\begin{equation}
    \bm\theta(\bm\alpha)
      =\bm\theta_0+\operatorname{Concat}_{\ell=0}^{L-1}
        \left(\sum_{k=1}^{K}\alpha_{\ell k}\Delta\bm\theta_{\ell}^{k}\right),
    \label{eq:direct_expert_merge}
\end{equation}
where $L$ is the number of parameter regions, $\Delta\bm\theta_\ell^k$ is expert $k$'s update block in region $\ell$, and $\alpha_{\ell k}$ is its weight.
Concatenation follows the original parameter order.
The vector $\bm\alpha$ collects these coefficients.
Given $\mathcal C$, we seek coefficients that preserve the complementary multimodal capabilities represented by the experts in one model shared across tasks.

\subsection{Framework Overview}
\label{sec:framework_overview}
\begin{figure}[t]
\centering
\includegraphics[width=\textwidth]{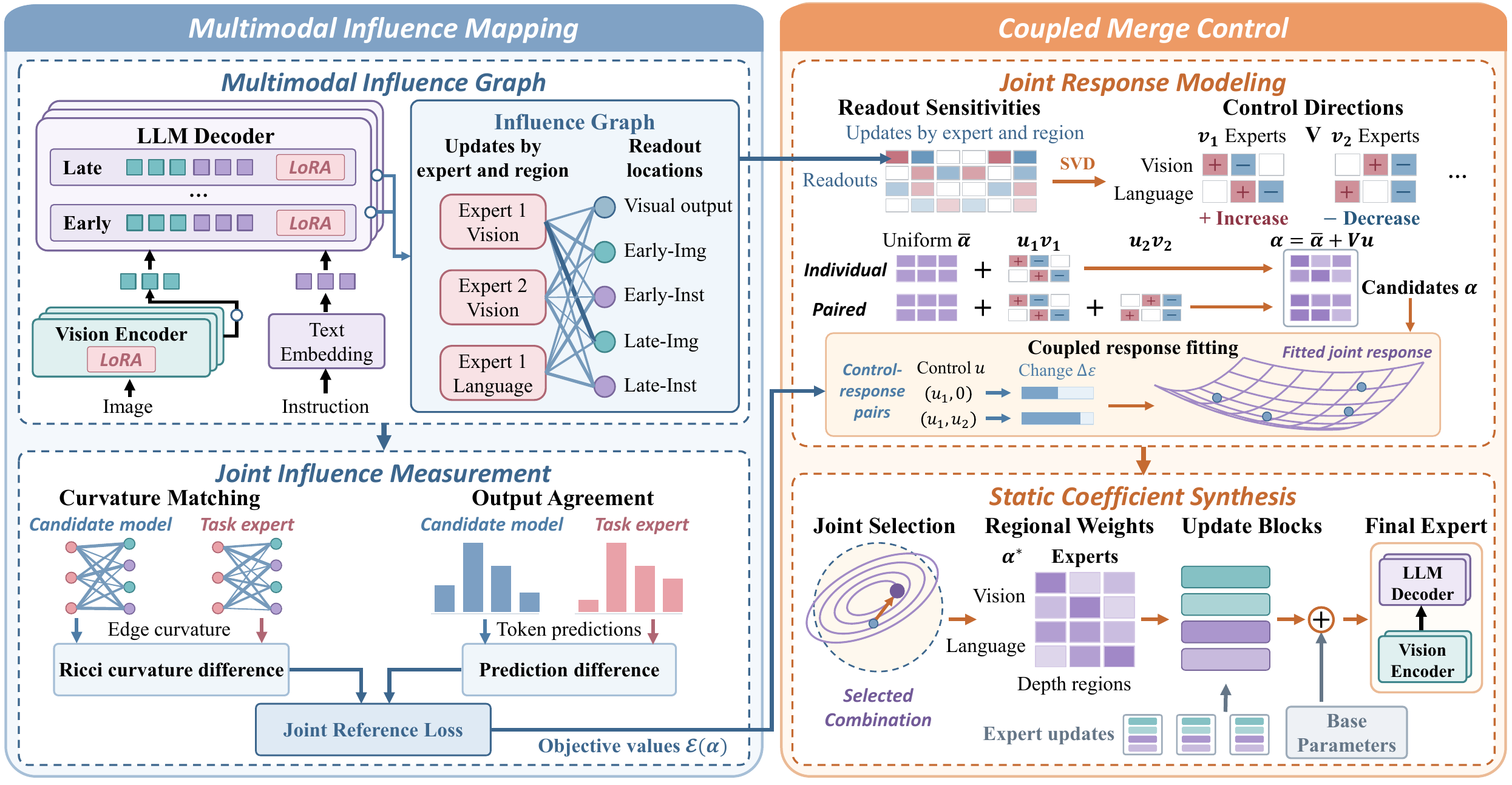}
\caption{Overview of \modelname. \moduleA~maps multimodal expert influence, and \moduleB~coordinates regional coefficients using the resulting reference objective to synthesize one shared model.}
\label{fig:Framework}
\end{figure}

Figure~\ref{fig:Framework} illustrates \modelname, which combines multimodal influence mapping (\moduleA) with coupled merge control (\moduleB).
\moduleA~constructs influence graphs from original readouts and responses to image and instruction changes, tracing regional update effects across depth and token positions.
Matching task-expert curvature and predictions defines the reference objective.
\moduleB~uses measured responses to identify a compact coefficient subspace and fits a quadratic model from individual and paired coefficient variations.
It jointly optimizes these coefficients around a uniform merge and accepts the proposal only if it improves the measured objective.
The selected weights combine regional expert updates into one shared model with fixed inference coefficients.

\subsection{Multimodal Influence Mapping}
\label{sec:influence_mapping}

\paragraph{Multimodal Influence Graph.}
Visual influence varies across depth and token positions~\citep{CrossModalFlow}, motivating a source-specific account of how expert updates alter image- and instruction-dependent computation.
The $L$ disjoint parameter regions are ordered along the visual and language pathways, with boundaries and readout positions fixed throughout coefficient selection.
The regional coefficients in Eq.~(\ref{eq:direct_expert_merge}) are shared across inputs and collected in $\bm\alpha\in\mathbb R^{LK}$.
The uniform reference $\bar{\bm\alpha}$ has every entry equal to $1/K$.
Let $\bm h_r(\bm v,\bm q,\bm\alpha)$ denote a pooled hidden readout, where $r$ specifies a region boundary and the group of token positions pooled there.
The readouts cover visual representations and language image-token and instruction-token positions carrying fused information~\citep{FGTracer}.
For each task $t$, a fixed seeded permutation forms disjoint calibration pairs $(i,j)$ with $i\ne j$.
Both pair orders are reused throughout coefficient selection, and any unpaired input still contributes to original readouts and output agreement.
For a fixed pair $(i,j)$, suppressing task superscripts, define the image-change and instruction-change responses separately:
\begin{equation}
    \bm h^v_{ij,r}(\bm\alpha)
      =\bigl(\bm h_r(\bm v_i,\bm q_i,\bm\alpha)
             -\bm h_r(\bm v_j,\bm q_i,\bm\alpha)\bigr),
    \quad
    \bm h^q_{ij,r}(\bm\alpha)
      =\bigl(\bm h_r(\bm v_i,\bm q_i,\bm\alpha)
             -\bm h_r(\bm v_i,\bm q_j,\bm\alpha)\bigr),
    \label{eq:conditional_readout}
\end{equation}
Here $\bm h^v_{ij,r}$ varies the image at a fixed instruction, whereas $\bm h^q_{ij,r}$ varies the instruction at a fixed image.
Both comparisons hold the readout location $r$ and coefficients $\bm\alpha$ fixed.
We retain the original representation as $\bm h^0_{i,r}(\bm\alpha)=\bm h_r(\bm v_i,\bm q_i,\bm\alpha)$ and index these three readout channels by $c\in\{0,v,q\}$.
Channel $0$ is always retained.
Channel $v$ or $q$ is retained when at least one fixed pair changes that modality, with this choice fixed throughout coefficient selection.
Identical instructions give $\bm h^q_{ij,r}=\bm0$ for every coefficient configuration, while the image-change response remains available.
For each available channel, a bipartite graph links expert--region sources to observations labeled by depth and token-position group.
An edge $(\ell,k,r)$ connects expert $k$ in region $\ell$ to observation $r$ reachable for channel $c$ in the forward computation.
The architecture and attention mask fix these connections, excluding observations structurally independent of the changed modality.
Its task-specific influence measures the sensitivity of the corresponding readout channel:
\begin{equation}
    w_{\ell k,r}^{t,c}(\bm\alpha)
      =\mathbb E
        \left[
          \left\|\frac{\partial\bm h^c_r(\bm\alpha)}
                           {\partial\alpha_{\ell k}}\right\|_2^2
        \right],
    \label{eq:influence_edges}
\end{equation}
where sample indices are suppressed in $\bm h^c_r$, and $\|\cdot\|_2$ is the Euclidean norm.
The expectation averages uniformly over all original inputs of task $t$ for $c=0$ and over its fixed ordered pairs for $c=v,q$.
The separate graphs preserve the distinction between the three sensitivities.
To characterize the allocation of influence around each node, incident weights are normalized into a probability distribution.
For a source, neighbor masses describe relative influence across observations, while for an observation they describe relative sensitivity to expert--region updates.
For nodes with positive total incident weight, suppressing task, channel, and coefficient arguments, we retain $m_{\ell k}(\ell,k)=m_r(r)=1/2$ at the source and observation nodes themselves and distribute the remaining mass among their architectural neighbors:
\begin{equation}
    m_{\ell k}(r)=\frac{w_{\ell k,r}}{2\sum_{r'}w_{\ell k,r'}},
    \qquad
    m_r(\ell,k)=\frac{w_{\ell k,r}}{2\sum_{\ell',k'}w_{\ell'k',r}}.
    \label{eq:graph_neighborhood}
\end{equation}
The sums range over architecturally connected observations and sources, respectively.
Other nodes receive zero probability mass.
A node with zero incident influence places all its probability mass at itself.
The equal self and neighbor masses retain endpoint identity and relative influence allocation for comparison across coefficient configurations.

\paragraph{Joint Influence Measurement.}
Joint updates change downstream computation~\citep{ChainOfMerges}, so selecting their coefficients requires characterizing their combined effect on multimodal influence.
For edge $(\ell,k,r)$ in channel $c$, let $m_{\ell k}$ and $m_r$ denote the endpoint neighborhood distributions defined above, with task, channel, and coefficient arguments suppressed.
Distances use the fixed shortest-path metric on the channel's undirected architectural support with unit edge lengths.
Since an edge's endpoints are at unit distance, its Ollivier--Ricci curvature~\citep{Ollivier2009} is:
\begin{equation}
    \kappa_{\ell k,r}^{t,c}(\bm\alpha)
      =1-W_1(m_{\ell k},m_r),
    \label{eq:ricci_curvature}
\end{equation}
where $W_1$ is the minimum cost of transporting one endpoint distribution to the other using this graph distance.
For a fixed graph support, changes in curvature summarize changes in the relative influence patterns around each source--observation pair.
The task-$t$ expert corresponds to coefficients equal to one for expert $t$ and zero for the other experts.
We evaluate its channel graphs on $\mathcal C_t$ using the same fixed pairs and record $\kappa_{\ell k,r}^{t,c,\star}$ as the task-specific curvature target.
For each original input, let $\bm p_i^t$ and $\bm p_i(\bm\alpha)$ denote expert and merged next-token distributions on fixed continuation prefixes generated by expert $t$.
Combining curvature matching with expert-output agreement gives:
\begin{equation}
    \mathcal E(\bm\alpha)
      =\mathbb E_{t,i}
        D_{\mathrm{KL}}\left(\bm p_i^t\,\|\,\bm p_i(\bm\alpha)\right)
       +\lambda\,\mathbb E_{t,c,\ell,k,r}
        \left[\bigl(\kappa_{\ell k,r}^{t,c}(\bm\alpha)
                    -\kappa_{\ell k,r}^{t,c,\star}\bigr)^2\right],
    \label{eq:geometric_reference}
\end{equation}
where $D_{\mathrm{KL}}$ is Kullback--Leibler divergence averaged over continuation positions and $\lambda\geq0$ weights the geometric reference.
Both expectations assign equal weight to tasks.
Within each task, the first weights original inputs equally, and the second weights available channels equally and architecturally valid edges equally within each channel.
Crossed inputs supply response measurements without answer targets.
At each queried coefficient configuration $\bm\alpha$, this module computes the readout derivatives and Eq.~(\ref{eq:geometric_reference}) on $\mathcal C$.
Curvature matching constrains the relative organization of influence across regions and modalities, while output agreement penalizes deviations from the expert's next-token distributions on the original calibration inputs.
Optimizing both terms links representation-level influence preservation to the predictive behavior needed for each task.

\subsection{Coupled Merge Control}
\label{sec:merge_control}

\paragraph{Joint Response Modeling.}
Selecting regional weights requires modeling how expert updates jointly affect the reference objective, since the effect of one update can depend on the others.
Directly modeling these interactions over all $LK$ coefficients is costly, so we construct a compact control subspace from measured multimodal responses and fit a local quadratic model within it.
To select this subspace, we stack the readout derivatives returned by \moduleA~at $\bar{\bm\alpha}$ into a response matrix, with one column per expert--region source and one row per scalar readout component across tasks, samples, channels, and readout locations.
The squared stacking weights assign equal mass to tasks, then to available channels, and uniform mass to samples and readout locations within each channel.
If the response matrix is zero, we retain the uniform merge.
For a nonzero matrix, let $\sigma_1\geq\sigma_2\geq\cdots$ denote its singular values.
We select the control dimension $d$ based on the stable rank $\sum_a\sigma_a^2/\sigma_1^2$, which reflects how response energy is distributed across singular directions.
The leading $d$ right singular vectors form an orthonormal basis $\bm V\in\mathbb R^{LK\times d}$, with each column specifying a coordinated adjustment of expert weights across regions.
The control vector $\bm u\in\mathbb R^d$ adjusts these directions within radius $R>0$ of the uniform reference:
\begin{equation}
    \bm\alpha=\bar{\bm\alpha}+\bm V\bm u,
    \qquad \|\bm u\|_2\leq R.
    \label{eq:control_coordinates}
\end{equation}
Orthonormality makes $R$ a bound on the aggregate coefficient displacement from $\bar{\bm\alpha}$.
Within this neighborhood, a quadratic model approximates the objective change relative to $\bar{\bm\alpha}$:
\begin{equation}
    \widehat{\mathcal E}(\bm u)
      =\bm g^\top\bm u+\tfrac12\bm u^\top\bm H\bm u,
    \label{eq:identified_control_model}
\end{equation}
where $\bm g\in\mathbb R^d$ represents linear trends and the symmetric matrix $\bm H\in\mathbb R^{d\times d}$ represents quadratic effects within and between control directions.
To fit these terms, we use \moduleA~to evaluate Eq.~(\ref{eq:geometric_reference}) under a fixed set of symmetric coefficient variations.
With $\bm e_a$ denoting a coordinate unit vector, we use signed individual variations $\bm u=\pm R\bm e_a$ for $1\leq a\leq d$.
We also use paired variations $\bm u=R(\pm\bm e_a\pm\bm e_b)/\sqrt2$ for $1\leq a<b\leq d$, with all independent sign choices.
Least-squares fitting then identifies the model coefficients:
\begin{equation}
    (\bm g,\bm H)\in
      \argmin_{\bm g,\,\bm H=\bm H^\top}
      \mathbb E_{\bm u}\left[
        \left(\mathcal E(\bar{\bm\alpha}+\bm V\bm u)
          -\mathcal E(\bar{\bm\alpha})
          -\bm g^\top\bm u-\tfrac12\bm u^\top\bm H\bm u\right)^2
      \right],
    \label{eq:control_identification}
\end{equation}
where $\mathbb E_{\bm u}$ averages over the evaluated coefficient configurations.
Individual variations identify linear trends and quadratic terms within each direction, while paired variations identify interactions between directions through the off-diagonal entries of $\bm H$.
The interaction terms let the controller evaluate each adjustment in the context of the others, helping identify expert combinations that work together across regions to preserve multimodal influence.

\paragraph{Static Coefficient Synthesis.}
Because the fitted model is only a local approximation, a predicted improvement may not reduce the measured fusion objective.
We therefore limit coefficient changes to the prescribed neighborhood and check the resulting proposal against the uniform reference.
We first obtain a joint coefficient proposal by minimizing the fitted objective:
\begin{equation}
    \bm u^\star\in\argmin_{\|\bm u\|_2\leq R}
      \widehat{\mathcal E}(\bm u),
    \label{eq:joint_control}
\end{equation}
The reference $\bm u=\bm0$ is feasible and predicts zero objective change.
We solve Eq.~(\ref{eq:joint_control}) using an eigendecomposition of $\bm H$ and the spectral trust-region conditions~\citep{MoreSorensen1983}:
\begin{equation}
    (\bm H+\mu\bm I_d)\bm u^\star=-\bm g,
    \qquad \bm H+\mu\bm I_d\succeq\bm0,
    \label{eq:control_solution}
\end{equation}
where $\mu\geq0$ is a spectral shift satisfying $\mu(\|\bm u^\star\|_2-R)=0$ and $\succeq\bm0$ denotes positive semidefiniteness.
An interior solution has $\mu=0$, while a boundary solution uses a shift consistent with $\|\bm u^\star\|_2=R$.
In the eigenbasis, the boundary search reduces to a scalar norm condition, with a nullspace component included when required by a singular boundary solution.
Together with feasibility, Eq.~(\ref{eq:control_solution}) characterizes a global minimizer of the fixed quadratic model over the prescribed ball.
To verify the predicted improvement, we evaluate the proposal using Eq.~(\ref{eq:geometric_reference}) on the same calibration data and expert targets.
The final coefficients are $\bm\alpha^\star=\bar{\bm\alpha}+\bm V\bm u^\star$ if $\mathcal E(\bar{\bm\alpha}+\bm V\bm u^\star)<\mathcal E(\bar{\bm\alpha})-\tau$, and $\bm\alpha^\star=\bar{\bm\alpha}$ otherwise, where $\tau\geq0$ is a fixed numerical comparison tolerance.
This check requires no iterative refitting of the response model.
The selected coefficients $\bm\alpha^\star$ synthesize the shared model:
\begin{equation}
    \bm\theta^\star
      =\bm\theta_0+\operatorname{Concat}_{\ell=0}^{L-1}
        \left(\sum_{k=1}^{K}
          \alpha_{\ell k}^\star\Delta\bm\theta_\ell^k\right),
    \label{eq:final_expert}
\end{equation}
The regional blocks are concatenated in the original parameter order to obtain $f(\cdot,\bm\theta^\star)$.
By selecting regional coefficients jointly, \moduleB~accounts for how updates in one part of the model affect the inputs seen by later updates.
This coordination helps reduce interference across layers and modalities and retain the complementary capabilities of different experts.

\section{Experiments}
\label{sec:experiments}

\subsection{Experimental Setups}
\label{sec:experimental_setups}

\paragraph{Datasets.}
Following RobustMerge~\citep{RobustMerge}, we adopt MM-MergeBench with eight seen datasets: ScienceQA~\citep{ScienceQA}, ImageNet~\citep{ImageNet}, VQAv2~\citep{VQAv2}, REC-COCO~\citep{ReferItGame,RefCOCOg}, OCR-VQA~\citep{OCRVQA}, VizWiz-Caption~\citep{VizWizCaptions}, Flickr30k~\citep{Flickr30k}, and IconQA~\citep{IconQA}.
These datasets cover visual question answering, referring expression comprehension, image classification, and image captioning.
Generalization is evaluated on four unseen datasets: A-OKVQA~\citep{AOKVQA}, ImageNet-R~\citep{ImageNetR}, Screen2Words~\citep{Screen2Words}, and TabMWP~\citep{TabMWP}.
The seen-task evaluation measures how well one merged model retains the capabilities of the eight adapted experts.
Unseen tasks evaluate whether the same coefficients transfer to datasets outside expert adaptation, testing whether the consolidated capabilities remain useful under a change in task distribution.

\paragraph{Baselines.}
To evaluate the effectiveness of \modelname, we compare it with the following representative baselines.
We first consider two reference settings:
(1) Zero-shot, which directly evaluates the pretrained model,
and (2) Individual, which evaluates each task expert on its corresponding task.
For single-modality merging, we include
(3) TA~\citep{TaskArithmetic}, which adds the summed expert updates to the base model with a shared scaling coefficient,
(4) TIES~\citep{TIESMerging}, which trims small updates, resolves sign conflicts, and averages the remaining updates with consistent signs,
and (5) DC-Merge~\citep{DCMerge}, which smooths singular values and aligns task vectors in a shared orthogonal subspace before aggregation.
For cross-layer merging, we compare
(6) Chain of Merges~\citep{ChainOfMerges}, which merges layers sequentially while refreshing activation statistics,
and (7) RegMean++~\citep{RegMean}, which incorporates intra-layer and cross-layer dependencies into regression-based merging.
For multimodal merging, we include
(8) RobustMerge~\citep{RobustMerge}, which prunes and rescales low-rank parameters and normalizes task contributions to preserve direction robustness,
and (9) OptMerge~\citep{OptMerge}, which denoises task vectors and optimizes the merged update through an objective over their interactions.

\paragraph{Implementation Details.}
We use Qwen3.5-4B~\citep{Qwen35} and Qwen3-VL-8B-Instruct~\citep{Qwen3VL} as the base models and train one LoRA expert for each of the eight seen tasks using Unsloth~\citep{Unsloth}.
LoRA adapters are inserted into the attention and MLP modules of both the vision encoder and the language backbone, and the two parts are optimized jointly on each task's image--instruction examples while the pretrained weights remain frozen.
The LoRA rank and scaling factor are both 16.
Experts train for one epoch with AdamW, learning rate $2\times10^{-4}$, linear decay, warmup ratio 0.03, weight decay 0.001, and a 2,048-token sequence limit.

\subsection{Performance Evaluation}
\label{sec:performance_evaluation}

\begin{table}[t]
    \centering
    \caption{Performance comparison on Qwen3.5-4B.}
    \label{tab:main_results_4b}
    \setlength{\tabcolsep}{3.2pt}
    \renewcommand{\arraystretch}{1.15}
    \resizebox{\textwidth}{!}{%
    \begin{tabular}{l*{14}{c}}
        \toprule
        & \multicolumn{9}{c}{Seen Tasks} & \multicolumn{5}{c}{Unseen Tasks} \\
        \cmidrule(lr){2-10}\cmidrule(lr){11-15}
        Method & SciQA & Image & VQA & REC & OCR & VizWiz & Flickr & IconQA & Avg. & AVQA & Image-R & S2W & TabMWP & Avg. \\
        \midrule
        Zero-shot & 67.13 & 38.81 & 72.00 & 76.15 & 70.58 & 23.74 & 25.65 & 25.86 & 49.99 & 78.69 & 87.27 & 5.35 & 64.39 & 58.92 \\
        Individual & 97.92 & 97.47 & 74.85 & 91.53 & 71.51 & 72.80 & 60.44 & 99.15 & 83.21 & -- & -- & -- & -- & -- \\
        \midrule
        TA & 94.99 & 59.03 & 72.52 & 73.48 & 68.78 & 52.21 & \underline{58.11} & 79.79 & 69.86 & 86.20 & \underline{90.95} & \underline{15.95} & 69.95 & \underline{65.76} \\
        TIES & 94.15 & 56.36 & \underline{72.71} & 74.22 & \underline{69.19} & 52.42 & 57.80 & 70.39 & 68.41 & 83.67 & \textbf{90.97} & 15.58 & 69.57 & 64.95 \\
        DC-Merge & 95.29 & 59.03 & 72.52 & 74.55 & 68.46 & 52.42 & 57.95 & 79.79 & 70.00 & 86.20 & 90.10 & 15.59 & 69.95 & 65.46 \\
        \midrule
        CoM & 95.29 & 63.21 & 72.54 & 74.62 & 68.46 & 52.59 & 58.04 & 79.25 & 70.50 & 84.72 & 90.10 & 15.59 & \underline{71.82} & 65.56 \\
        RegMean++ & 95.84 & 85.86 & 69.92 & \underline{75.55} & 67.41 & \underline{53.16} & 57.95 & \underline{94.85} & \underline{75.07} & \underline{87.25} & 85.83 & 15.19 & 70.78 & 64.76 \\
        \midrule
        RobustMerge & \underline{96.03} & 48.18 & 72.15 & 74.55 & 66.75 & 50.76 & 57.59 & 92.15 & 69.77 & 87.07 & 88.83 & 15.75 & 67.85 & 64.88 \\
        OptMerge & 93.80 & \textbf{88.46} & 63.25 & 38.70 & 57.04 & 48.76 & 56.50 & 47.83 & 61.79 & 58.08 & 77.86 & 14.92 & 53.86 & 51.18 \\
        \midrule
        \textbf{\modelname} & \textbf{97.12} & \underline{88.08} & \textbf{73.86} & \textbf{83.42} & \textbf{69.93} & \textbf{58.68} & \textbf{59.84} & \textbf{97.07} & \textbf{78.50} & \textbf{88.47} & 90.57 & \textbf{18.72} & \textbf{74.64} & \textbf{68.10} \\
        \bottomrule
    \end{tabular}%
    }
\end{table}

\begin{table}[t]
    \centering
    \caption{Performance comparison on Qwen3-VL-8B.}
    \label{tab:main_results_8b}
    \setlength{\tabcolsep}{3.2pt}
    \renewcommand{\arraystretch}{1.15}
    \resizebox{\textwidth}{!}{%
    \begin{tabular}{l*{14}{c}}
        \toprule
        & \multicolumn{9}{c}{Seen Tasks} & \multicolumn{5}{c}{Unseen Tasks} \\
        \cmidrule(lr){2-10}\cmidrule(lr){11-15}
        Method & SciQA & Image & VQA & REC & OCR & VizWiz & Flickr & IconQA & Avg. & AVQA & Image-R & S2W & TabMWP & Avg. \\
        \midrule
        Zero-shot & 94.40 & 35.03 & 72.34 & 77.32 & 69.61 & 27.26 & 28.66 & 44.20 & 56.10 & 79.65 & 90.48 & 5.87 & 60.40 & 59.10 \\
        Individual & 95.79 & 94.69 & 74.45 & 89.33 & 74.32 & 69.68 & 59.54 & 90.94 & 81.09 & -- & -- & -- & -- & -- \\
        \midrule
        TA & 95.19 & 35.07 & 72.87 & 77.08 & 72.43 & 48.78 & 56.86 & 72.29 & 66.32 & 87.16 & \textbf{90.67} & 9.41 & 61.67 & 62.23 \\
        TIES & 96.13 & 54.77 & 70.65 & 84.07 & 70.24 & 51.12 & \underline{58.18} & 82.49 & 70.96 & 86.72 & 89.60 & 14.04 & 63.14 & 63.38 \\
        DC-Merge & \textbf{96.58} & 72.08 & 70.52 & \underline{86.55} & 73.25 & \underline{52.76} & 57.67 & \underline{86.94} & 74.54 & 87.34 & 89.95 & \underline{14.41} & 66.56 & \underline{64.56} \\
        \midrule
        CoM & 95.98 & 46.85 & 72.67 & 74.13 & 72.59 & 51.38 & 57.89 & 83.40 & 69.36 & 87.51 & 90.07 & 13.03 & 64.10 & 63.68 \\
        RegMean++ & \underline{96.43} & \underline{77.49} & \underline{73.03} & 83.92 & \underline{73.27} & 52.33 & 57.77 & 86.45 & \underline{75.08} & \underline{87.95} & 90.33 & 12.35 & \underline{67.55} & 64.55 \\
        \midrule
        RobustMerge & 95.14 & 34.85 & 72.68 & 69.90 & 71.61 & 49.83 & 58.17 & 80.78 & 66.62 & \underline{87.95} & 90.22 & 14.35 & 64.05 & 64.14 \\
        OptMerge & 96.23 & 58.77 & 70.76 & 84.15 & 70.77 & 50.25 & 57.78 & 82.16 & 71.36 & 87.60 & 89.75 & 13.33 & 61.75 & 63.11 \\
        \midrule
        \textbf{\modelname} & 96.38 & \textbf{82.63} & \textbf{73.58} & \textbf{87.42} & \textbf{73.46} & \textbf{56.83} & \textbf{59.37} & \textbf{90.33} & \textbf{77.50} & \textbf{88.65} & \underline{90.53} & \textbf{16.84} & \textbf{69.98} & \textbf{66.50} \\
        \bottomrule
    \end{tabular}%
    }
\end{table}

\paragraph{Main Results.}
Tables~\ref{tab:main_results_4b} and~\ref{tab:main_results_8b} compare performance on seen and unseen tasks using Qwen3.5-4B and Qwen3-VL-8B, respectively.
The results yield four main observations.
(1) Traditional single-modality merging methods struggle to balance heterogeneous multimodal capabilities.
Task Arithmetic, TIES, and DC-Merge determine update compatibility through parameter aggregation, sign consistency, or subspace alignment, leaving the role of each update in visual--language computation implicit.
Parameter-compatible updates can still alter the representations through which visual and textual information interact.
The resulting trade-offs are evident on the 8B model, where TIES gains 19.70 points on ImageNet over Task Arithmetic while losing 2.22 points on VQAv2.
(2) Existing multimodal merging methods remain limited in preserving the propagation of expert capabilities across depth.
RobustMerge stabilizes low-rank update directions, and OptMerge optimizes interactions among task vectors within individual layers.
These parameter-space objectives provide indirect control over how a visual update changes the input encountered by later language updates.
Their lower aggregate performance is consistent with this gap between local update compatibility and preservation of multimodal influence throughout the model.
(3) Cross-layer methods improve knowledge consolidation, with remaining limitations in multimodal transfer.
Chain of Merges refreshes activation statistics sequentially, and RegMean++ models correlations across layers, accounting for representation changes induced by preceding updates.
Their objectives emphasize activation agreement and cross-layer correlations, leaving source-specific image--instruction effects implicit.
On the 4B model, RegMean++ improves the seen average over TIES by 6.66 points, while its unseen average remains 1.00 point below Task Arithmetic.
This uneven transfer supports extending cross-layer dependency modeling with explicit multimodal influence measurements when jointly selecting expert coefficients.
(4) \modelname~achieves the best average performance on both seen and unseen tasks across the two backbones.
On Qwen3.5-4B, it exceeds the strongest merging baselines by 3.43 points on seen tasks and 2.34 points on unseen tasks.
The corresponding gains on Qwen3-VL-8B are 2.42 and 1.94 points, respectively.
It also leads the merging methods on seven of eight seen tasks and three of four unseen tasks for each backbone.
These results support jointly selecting regional coefficients based on multimodal influence to preserve complementary expert capabilities in a shared model.

\begin{table}[t]
    \centering
    \caption{Ablation study.}
    \label{tab:ablation_study}
    \small
    \setlength{\tabcolsep}{4pt}
    \renewcommand{\arraystretch}{1.17}
    \begin{tabular*}{\linewidth}{@{\extracolsep{\fill}}lcccc@{}}
        \toprule
        & \multicolumn{2}{c}{Qwen3.5-4B} & \multicolumn{2}{c}{Qwen3-VL-8B} \\
        \cmidrule(lr){2-3}\cmidrule(lr){4-5}
        Variant & Seen & Unseen & Seen & Unseen \\
        \midrule
        w/o \moduleA & $76.12\pm0.39$ & $67.04\pm0.27$ & $76.03\pm0.43$ & $65.88\pm0.30$ \\
        \moduleA: w/o modality contrasts & $77.46\pm0.26$ & $67.87\pm0.21$ & $77.09\pm0.29$ & $66.29\pm0.24$ \\
        \midrule
        w/o \moduleB & $77.03\pm0.46$ & $67.58\pm0.32$ & $76.64\pm0.48$ & $66.26\pm0.34$ \\
        \moduleB: w/o directional coupling & $77.79\pm0.30$ & $67.71\pm0.20$ & $77.02\pm0.36$ & $66.38\pm0.19$ \\
        \midrule
        \textbf{\modelname~(Full)} & $\bm{78.50\pm0.24}$ & $\bm{68.10\pm0.18}$ & $\bm{77.50\pm0.31}$ & $\bm{66.50\pm0.22}$ \\
        \bottomrule
    \end{tabular*}
\end{table}

\paragraph{Ablation Study.}
Table~\ref{tab:ablation_study} presents ablation results organized by module, examining each module and its central design.
Results are reported as mean $\pm$ standard deviation.
All variants share the expert checkpoints, calibration inputs, coefficient radius, and budget of objective evaluations, with the control dimension matched whenever a basis is used.
(1) Effect of \moduleA.
The variant w/o \moduleA~uses output KL as its sole objective and a fixed random orthonormal basis for \moduleB, replacing the influence graphs and readout-response measurements.
The variant w/o modality contrasts retains the original readouts $\bm h^0$ and their influence graphs, removing the image-change and instruction-change channels from both curvature matching and response-based subspace selection.
Original readouts capture overall representation changes, while the two contrast channels distinguish update effects associated with changing the image or instruction at fixed observation locations.
Removing \moduleA~reduces seen averages by 2.38 and 1.47 points on 4B and 8B, respectively, compared with 1.04 and 0.41 points when removing only the contrasts.
These results support using explicit modality-dependent responses alongside the original representations to guide influence matching and coefficient-direction selection.
(2) Effect of \moduleB.
The variant w/o \moduleB~retains the full reference objective and replaces subspace selection and quadratic control with direct search over fixed-seed regional coefficient candidates sampled uniformly within the same radius, selecting the best candidate together with the uniform reference.
The variant w/o directional coupling retains the fitted linear terms and the diagonal entries of the quadratic matrix $\bm H$, setting its off-diagonal entries to zero before solving for the coefficients.
Its response basis, calibration evaluations, and proposal check remain fixed, isolating the contribution of interactions between control directions.
Each direction adjusts multiple regional weights, so these mixed terms describe how one coordinated adjustment changes the effect of another.
The unseen averages decrease by 0.52 and 0.24 points after removing \moduleB, compared with 0.39 and 0.12 points after removing only directional coupling.
These results support coordinated coefficient selection that accounts for interactions among regional updates, helping preserve complementary capabilities on unseen tasks.
(3) Overall comparison.
Across both backbones, the full configuration achieves the highest mean in all four evaluation settings, while removing either module lowers both seen and unseen performance.
Retaining \moduleA~but replacing \moduleB~with direct search reduces the seen averages by 1.47 and 0.86 points on 4B and 8B, respectively, suggesting that an informative fusion objective alone may be insufficient under a fixed evaluation budget.
Conversely, the larger drops after removing \moduleA~suggest that coefficient control depends on the objective and response measurements used to guide it.
These comparisons distinguish two complementary roles.
\moduleA~specifies the multimodal influence to preserve through its reference objective, while \moduleB~uses that objective to compare coordinated coefficient adjustments and select regional weights for the merged model.

\begin{figure}
    \centering
    \includegraphics[width=\linewidth]{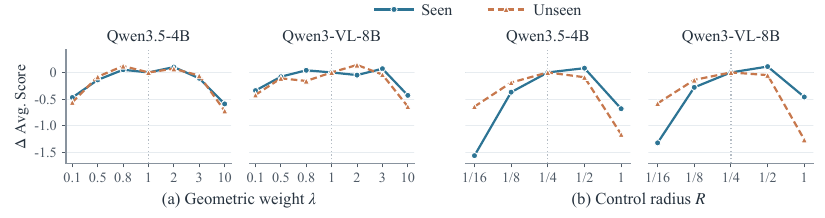}
    \caption{Sensitivity to (a) $\lambda$ and (b) $R$. $\Delta$ is the average-score change from $(\lambda,R)=(1,0.25)$ in score points. Settings are equally spaced, and dotted lines mark the default values.}
    \label{fig:hyperparameter_sensitivity}
\end{figure}

\paragraph{Hyperparameter Sensitivity.}
To examine the sensitivity of \modelname, Figure~\ref{fig:hyperparameter_sensitivity} compares changes in seen and unseen average scores relative to the shared default $(\lambda,R)=(1,0.25)$.
The scans vary $\lambda\in\{0.1,0.5,0.8,1,2,3,10\}$ and $R\in\{1/16,1/8,1/4,1/2,1\}$ separately, holding the other parameter fixed.
Each curve uses its corresponding default mean from the main tables as zero, allowing seen and unseen sensitivity to be compared on the same scale.
The sensitivity curves show a broad intermediate plateau for $\lambda$, with small local fluctuations.
Smaller weights retain expert-output agreement and the response subspace, while larger weights emphasize the experts' relative influence patterns.
Since these geometric targets are constructed from seen tasks, preserving their influence patterns and transferring to unseen inputs can favor different weights.
For $R$, smaller neighborhoods restrict coordinated coefficient adjustments, whereas larger neighborhoods expand both the feasible displacement and the configurations used to fit the response model.
The large-radius reference points represent accepted proposals with a greater decline on unseen tasks, illustrating a possible difference between calibration-objective improvement and transfer to new task distributions.
The acceptance check can also produce plateaus when multiple proposals return the uniform merge.

\section{Conclusion}
\label{sec:conclusion}

This paper presents \modelname, a framework for multimodal model merging that traces expert influence across depth and jointly coordinates regional updates.
Multimodal influence mapping characterizes how each update affects representations at different depths and token positions, using Ricci curvature and expert predictions to define a shared merging objective.
Coupled merge control models interactions among coefficient adjustments in a compact subspace and selects regional weights to synthesize one model shared across tasks.
The selected coefficients yield one static model that serves all tasks with the base inference architecture.
Across the two evaluated backbones, the merged model improves both seen and unseen averages over the compared merging methods, and the ablations support the contribution of modality-specific responses and interactions between control directions.
Future work will explore more efficient influence estimation and scalable coordination when merging larger collections of task experts.

\bibliography{iclr2027_conference}

@article{MaTS,
  title={Merging by Matching Models in Task Parameter Subspaces},
  author={Tam, Derek and Bansal, Mohit and Raffel, Colin},
  journal={Transactions on Machine Learning Research},
  year={2024},
  url={https://arxiv.org/abs/2312.04339}
}

@article{RegMean,
title={RegMean++: Enhancing Effectiveness and Generalization of Regression Mean for Model Merging},
author={The-Hai Nguyen and Dang Huu-Tien and Takeshi Suzuki and Le-Minh Nguyen},
journal={Transactions on Machine Learning Research},
issn={2835-8856},
year={2026},
url={https://openreview.net/forum?id=H5lDsSCS9i},
note={}
}

@inproceedings{LoRA,
title={Lo{RA}: Low-Rank Adaptation of Large Language Models},
author={Edward J Hu and yelong shen and Phillip Wallis and Zeyuan Allen-Zhu and Yuanzhi Li and Shean Wang and Lu Wang and Weizhu Chen},
booktitle={International Conference on Learning Representations},
year={2022},
url={https://openreview.net/forum?id=nZeVKeeFYf9}
}

@inproceedings{TaskArithmetic,
title={Editing models with task arithmetic},
author={Gabriel Ilharco and Marco Tulio Ribeiro and Mitchell Wortsman and Ludwig Schmidt and Hannaneh Hajishirzi and Ali Farhadi},
booktitle={The Eleventh International Conference on Learning Representations },
year={2023},
url={https://openreview.net/forum?id=6t0Kwf8-jrj}
}

@inproceedings{TIESMerging,
author = {Yadav, Prateek and Tam, Derek and Choshen, Leshem and Raffel, Colin and Bansal, Mohit},
booktitle = {Advances in Neural Information Processing Systems},
doi = {10.52202/075280-0310},
editor = {A. Oh and T. Naumann and A. Globerson and K. Saenko and M. Hardt and S. Levine},
pages = {7093--7115},
publisher = {Curran Associates, Inc.},
title = {TIES-Merging: Resolving Interference When Merging Models},
url = {https://proceedings.neurips.cc/paper_files/paper/2023/file/1644c9af28ab7916874f6fd6228a9bcf-Paper-Conference.pdf},
volume = {36},
year = {2023}
}

@inproceedings{ModelComposition,
title = "Model Composition for Multimodal Large Language Models",
author = "Chen, Chi  and
  Du, Yiyang  and
  Fang, Zheng  and
  Wang, Ziyue  and
  Luo, Fuwen  and
  Li, Peng  and
  Yan, Ming  and
  Zhang, Ji  and
  Huang, Fei  and
  Sun, Maosong  and
  Liu, Yang",
editor = "Ku, Lun-Wei  and
  Martins, Andre  and
  Srikumar, Vivek",
booktitle = "Proceedings of the 62nd Annual Meeting of the Association for Computational Linguistics (Volume 1: Long Papers)",
month = aug,
year = "2024",
address = "Bangkok, Thailand",
publisher = "Association for Computational Linguistics",
url = "https://aclanthology.org/2024.acl-long.606/",
doi = "10.18653/v1/2024.acl-long.606",
pages = "11246--11262"
}

@InProceedings{AdaMMS,
author    = {Du, Yiyang and Wang, Xiaochen and Chen, Chi and Ye, Jiabo and Wang, Yiru and Li, Peng and Yan, Ming and Zhang, Ji and Huang, Fei and Sui, Zhifang and Sun, Maosong and Liu, Yang},
title     = {AdaMMS: Model Merging for Heterogeneous Multimodal Large Language Models with Unsupervised Coefficient Optimization},
booktitle = {Proceedings of the IEEE/CVF Conference on Computer Vision and Pattern Recognition (CVPR)},
month     = {June},
year      = {2025},
pages     = {9413-9422}
}

@inproceedings{ExpertMerging,
author = {Zhang, Dengming and Ma, Xiaowen and Ni, Zhenliang and Wu, Zhenkai and Shu, Han and Jiang, Xin and Chen, Xinghao},
booktitle = {International Conference on Learning Representations},
editor = {C. Vondrick and B. Hariharan and C. Raffel and L. Pinto and D. Yang and A. Faust},
pages = {21780--21803},
title = {Expert Merging: Model Merging with Unsupervised Expert Alignment and Importance-Guided Layer Chunking},
url = {https://proceedings.iclr.cc/paper_files/paper/2026/file/24dbe8babeca9f372712bbf19dd3584e-Paper-Conference.pdf},
volume = {2026},
year = {2026}
}

@inproceedings{OptMerge,
author = {Wei, Yongxian and Cheng, Runxi and Jin, Weike and Yang, Enneng and Shen, Li and HOU, LU and Du, SiNan and Yuan, Chun and Cao, Xiaochun and Tao, Dacheng},
booktitle = {International Conference on Learning Representations},
editor = {C. Vondrick and B. Hariharan and C. Raffel and L. Pinto and D. Yang and A. Faust},
pages = {61571--61595},
title = {OptMerge: Unifying Multimodal LLM Capabilities and Modalities via Model Merging},
url = {https://proceedings.iclr.cc/paper_files/paper/2026/file/64741c1e80cc3bbed205b1c2b040dfa8-Paper-Conference.pdf},
volume = {2026},
year = {2026}
}

@InProceedings{ModelSoups,
  title = 	 {Model soups: averaging weights of multiple fine-tuned models improves accuracy without increasing inference time},
  author =       {Wortsman, Mitchell and Ilharco, Gabriel and Gadre, Samir Ya and Roelofs, Rebecca and Gontijo-Lopes, Raphael and Morcos, Ari S and Namkoong, Hongseok and Farhadi, Ali and Carmon, Yair and Kornblith, Simon and Schmidt, Ludwig},
  booktitle = 	 {Proceedings of the 39th International Conference on Machine Learning},
  pages = 	 {23965--23998},
  year = 	 {2022},
  editor = 	 {Chaudhuri, Kamalika and Jegelka, Stefanie and Song, Le and Szepesvari, Csaba and Niu, Gang and Sabato, Sivan},
  volume = 	 {162},
  series = 	 {Proceedings of Machine Learning Research},
  month = 	 {17--23 Jul},
  publisher =    {PMLR},
  url = 	 {https://proceedings.mlr.press/v162/wortsman22a.html},
}

@InProceedings{TSVMerging,
author    = {Gargiulo, Antonio Andrea and Crisostomi, Donato and Bucarelli, Maria Sofia and Scardapane, Simone and Silvestri, Fabrizio and Rodol\`a, Emanuele},
title     = {Task Singular Vectors: Reducing Task Interference in Model Merging},
booktitle = {Proceedings of the IEEE/CVF Conference on Computer Vision and Pattern Recognition (CVPR)},
month     = {June},
year      = {2025},
pages     = {18695-18705}
}

@inproceedings{RobustMerge,
author = {Zeng, Fanhu and Guo, Haiyang and Zhu, Fei and Shen, Li and Tang, Hao},
booktitle = {Advances in Neural Information Processing Systems},
doi = {10.52202/085713-2390},
editor = {D. Belgrave and C. Zhang and H. Lin and R. Pascanu and P. Koniusz and M. Ghassemi and N. Chen},
pages = {71071--71095},
publisher = {Curran Associates, Inc.},
title = {RobustMerge: Parameter-Efficient Model Merging for MLLMs with Direction Robustness},
url = {https://proceedings.neurips.cc/paper_files/paper/2025/file/67101f97dc23fcc10346091181fff6cb-Paper-Conference.pdf},
volume = {38, Main Conference},
year = {2025}
}

@inproceedings{AdaMerging,
author = {Yang, Enneng and Wang, Zhenyi and Shen, Li and Liu, Shiwei and Guo, Guibing and Wang, Xingwei and Tao, Dacheng},
booktitle = {International Conference on Learning Representations},
editor = {B. Kim and Y. Yue and S. Chaudhuri and K. Fragkiadaki and M. Khan and Y. Sun},
pages = {22743--22763},
title = {AdaMerging: Adaptive Model Merging for Multi-Task Learning},
url = {https://proceedings.iclr.cc/paper_files/paper/2024/file/62868cc2fc1eb5cdf321d05b4b88510c-Paper-Conference.pdf},
volume = {2024},
year = {2024}
}

@inproceedings{UQMerge,
title = "$\texttt{UQ-Merge}$: Uncertainty Guided Multimodal Large Language Model Merging",
author = "Qu, Huaizhi  and
  Zhao, Xinyu  and
  Peng, Jie  and
  Lee, Kwonjoon  and
  Dariush, Behzad  and
  Chen, Tianlong",
editor = "Che, Wanxiang  and
  Nabende, Joyce  and
  Shutova, Ekaterina  and
  Pilehvar, Mohammad Taher",
booktitle = "Findings of the Association for Computational Linguistics: ACL 2025",
month = jul,
year = "2025",
address = "Vienna, Austria",
publisher = "Association for Computational Linguistics",
url = "https://aclanthology.org/2025.findings-acl.73/",
doi = "10.18653/v1/2025.findings-acl.73",
pages = "1401--1417",
ISBN = "979-8-89176-256-5",
}

@InProceedings{CrossModalFlow,
author    = {Zhang, Zhi and Yadav, Srishti and Han, Fengze and Shutova, Ekaterina},
title     = {Cross-modal Information Flow in Multimodal Large Language Models},
booktitle = {Proceedings of the IEEE/CVF Conference on Computer Vision and Pattern Recognition (CVPR)},
month     = {June},
year      = {2025},
pages     = {19781-19791}
}

@misc{ChainOfMerges,
title={Rethinking Layer-wise Model Merging through Chain of Merges}, 
author={Pietro Buzzega and Riccardo Salami and Angelo Porrello and Simone Calderara},
year={2026},
eprint={2508.21421},
archivePrefix={arXiv},
primaryClass={cs.LG},
url={https://arxiv.org/abs/2508.21421}, 
}

@InProceedings{FGTracer,
author    = {Saporita, Alessia and Pipoli, Vittorio and Bolelli, Federico and Baraldi, Lorenzo and Acquaviva, Andrea and Ficarra, Elisa},
title     = {FG-TRACER: Tracing Information Flow in Multimodal Large Language Models in Free-Form Generation},
booktitle = {Proceedings of the IEEE/CVF Winter Conference on Applications of Computer Vision (WACV)},
month     = {March},
year      = {2026},
pages     = {7903-7912}
}

@InProceedings{DAREMerging,
title = 	 {Language Models are Super Mario: Absorbing Abilities from Homologous Models as a Free Lunch},
author =       {Yu, Le and Yu, Bowen and Yu, Haiyang and Huang, Fei and Li, Yongbin},
booktitle = 	 {Proceedings of the 41st International Conference on Machine Learning},
pages = 	 {57755--57775},
year = 	 {2024},
editor = 	 {Salakhutdinov, Ruslan and Kolter, Zico and Heller, Katherine and Weller, Adrian and Oliver, Nuria and Scarlett, Jonathan and Berkenkamp, Felix},
volume = 	 {235},
series = 	 {Proceedings of Machine Learning Research},
month = 	 {21--27 Jul},
publisher =    {PMLR},
url = 	 {https://proceedings.mlr.press/v235/yu24p.html},
}

@inproceedings{FisherMerging,
author = {Matena, Michael S and Raffel, Colin},
booktitle = {Advances in Neural Information Processing Systems},
doi = {10.52202/068431-1287},
editor = {S. Koyejo and S. Mohamed and A. Agarwal and D. Belgrave and K. Cho and A. Oh},
pages = {17703--17716},
publisher = {Curran Associates, Inc.},
title = {Merging Models with Fisher-Weighted Averaging},
url = {https://proceedings.neurips.cc/paper_files/paper/2022/file/70c26937fbf3d4600b69a129031b66ec-Paper-Conference.pdf},
volume = {35},
year = {2022}
}

@inproceedings{KnOTSMerging,
author = {Stoica, George and Ramesh, Pratik and Ecsedi, Boglarka and Choshen, Leshem and Hoffman, Judy},
booktitle = {International Conference on Learning Representations},
editor = {Y. Yue and A. Garg and N. Peng and F. Sha and R. Yu},
pages = {4501--4519},
title = {Model merging with SVD to tie the Knots},
url = {https://proceedings.iclr.cc/paper_files/paper/2025/file/0d4f8a5109c5083b5307fcd0bddae7a7-Paper-Conference.pdf},
volume = {2025},
year = {2025}
}

@inproceedings{RegMeanOriginal,
title={Dataless Knowledge Fusion by Merging Weights of Language Models},
author={Xisen Jin and Xiang Ren and Daniel Preotiuc-Pietro and Pengxiang Cheng},
booktitle={The Eleventh International Conference on Learning Representations },
year={2023},
url={https://openreview.net/forum?id=FCnohuR6AnM}
}

@InProceedings{Patchscopes,
title = 	 {Patchscopes: A Unifying Framework for Inspecting Hidden Representations of Language Models},
author =       {Ghandeharioun, Asma and Caciularu, Avi and Pearce, Adam and Dixon, Lucas and Geva, Mor},
booktitle = 	 {Proceedings of the 41st International Conference on Machine Learning},
pages = 	 {15466--15490},
year = 	 {2024},
editor = 	 {Salakhutdinov, Ruslan and Kolter, Zico and Heller, Katherine and Weller, Adrian and Oliver, Nuria and Scarlett, Jonathan and Berkenkamp, Felix},
volume = 	 {235},
series = 	 {Proceedings of Machine Learning Research},
month = 	 {21--27 Jul},
publisher =    {PMLR},
url = 	 {https://proceedings.mlr.press/v235/ghandeharioun24a.html},
}

@inproceedings{DCMerge,
  author={Zhang, Han-Chen and Zhou, Zi-Hao and Luo, Mao-Lin and Di, Shimin and Zhang, Min-Ling and Wei, Tong},
  title={{DC-Merge}: Improving Model Merging with Directional Consistency},
  booktitle={Proceedings of the IEEE/CVF Conference on Computer Vision and Pattern Recognition (CVPR)},
  month=jun,
  year={2026},
  pages={22248--22258},
  url={https://openaccess.thecvf.com/content/CVPR2026/html/Zhang_DC-Merge_Improving_Model_Merging_with_Directional_Consistency_CVPR_2026_paper.html}
}

@misc{OutputSpaceProjection,
  author={Evans, Bethan and Etheridge, Benjamin and Roberts, Stephen and Tanner, Jared},
  title={Model Merging by Output-Space Projection},
  year={2026},
  howpublished={arXiv preprint arXiv:2605.29101},
  eprint={2605.29101},
  archivePrefix={arXiv},
  primaryClass={cs.LG},
  url={https://arxiv.org/abs/2605.29101}
}

@article{Ollivier2009,
  author={Ollivier, Yann},
  title={Ricci curvature of {Markov} chains on metric spaces},
  journal={Journal of Functional Analysis},
  volume={256},
  number={3},
  pages={810--864},
  year={2009},
  doi={10.1016/j.jfa.2008.11.001},
  url={https://doi.org/10.1016/j.jfa.2008.11.001}
}

@article{MoreSorensen1983,
  author={Mor{\'e}, Jorge J. and Sorensen, D. C.},
  title={Computing a Trust Region Step},
  journal={SIAM Journal on Scientific and Statistical Computing},
  volume={4},
  number={3},
  pages={553--572},
  year={1983},
  doi={10.1137/0904038},
  url={https://doi.org/10.1137/0904038}
}

@inproceedings{ScienceQA,
  author={Lu, Pan and Mishra, Swaroop and Xia, Tanglin and Qiu, Liang and Chang, Kai-Wei and Zhu, Song-Chun and Tafjord, Oyvind and Clark, Peter and Kalyan, Ashwin},
  title={Learn to Explain: Multimodal Reasoning via Thought Chains for Science Question Answering},
  booktitle={Advances in Neural Information Processing Systems},
  volume={35},
  pages={2507--2521},
  year={2022},
  url={https://arxiv.org/abs/2209.09513}
}

@inproceedings{ImageNet,
  author={Deng, Jia and Dong, Wei and Socher, Richard and Li, Li-Jia and Li, Kai and Fei-Fei, Li},
  title={{ImageNet}: A Large-Scale Hierarchical Image Database},
  booktitle={Proceedings of the IEEE Conference on Computer Vision and Pattern Recognition (CVPR)},
  pages={248--255},
  year={2009},
  doi={10.1109/CVPR.2009.5206848}
}

@inproceedings{VQAv2,
  author={Goyal, Yash and Khot, Tejas and Summers-Stay, Douglas and Batra, Dhruv and Parikh, Devi},
  title={Making the {V} in {VQA} Matter: Elevating the Role of Image Understanding in Visual Question Answering},
  booktitle={Proceedings of the IEEE Conference on Computer Vision and Pattern Recognition (CVPR)},
  pages={6904--6913},
  year={2017},
  url={https://arxiv.org/abs/1612.00837}
}

@inproceedings{ReferItGame,
  author={Kazemzadeh, Sahar and Ordonez, Vicente and Matten, Mark and Berg, Tamara},
  title={{ReferItGame}: Referring to Objects in Photographs of Natural Scenes},
  booktitle={Proceedings of the 2014 Conference on Empirical Methods in Natural Language Processing (EMNLP)},
  pages={787--798},
  year={2014},
  publisher={Association for Computational Linguistics},
  doi={10.3115/v1/D14-1086},
  url={https://aclanthology.org/D14-1086/}
}

@inproceedings{RefCOCOg,
  author={Mao, Junhua and Huang, Jonathan and Toshev, Alexander and Camburu, Oana and Yuille, Alan L. and Murphy, Kevin},
  title={Generation and Comprehension of Unambiguous Object Descriptions},
  booktitle={Proceedings of the IEEE Conference on Computer Vision and Pattern Recognition (CVPR)},
  pages={11--20},
  year={2016},
  url={https://arxiv.org/abs/1511.02283}
}

@inproceedings{OCRVQA,
  author={Mishra, Anand and Shekhar, Shashank and Singh, Ajeet Kumar and Chakraborty, Anirban},
  title={{OCR-VQA}: Visual Question Answering by Reading Text in Images},
  booktitle={International Conference on Document Analysis and Recognition (ICDAR)},
  year={2019},
  url={https://ocr-vqa.github.io/}
}

@inproceedings{VizWizCaptions,
  author={Gurari, Danna and Zhao, Yinan and Zhang, Meng and Bhattacharya, Nilavra},
  title={Captioning Images Taken by People Who Are Blind},
  booktitle={European Conference on Computer Vision (ECCV)},
  year={2020},
  url={https://arxiv.org/abs/2002.08565}
}

@article{Flickr30k,
  author={Plummer, Bryan A. and Wang, Liwei and Cervantes, Chris M. and Caicedo, Juan C. and Hockenmaier, Julia and Lazebnik, Svetlana},
  title={{Flickr30k} Entities: Collecting Region-to-Phrase Correspondences for Richer Image-to-Sentence Models},
  journal={International Journal of Computer Vision},
  volume={123},
  number={1},
  pages={74--93},
  year={2017},
  doi={10.1007/s11263-016-0965-7}
}

@inproceedings{IconQA,
  author={Lu, Pan and Qiu, Liang and Chen, Jiaqi and Xia, Tony and Zhao, Yizhou and Zhang, Wei and Yu, Zhou and Liang, Xiaodan and Zhu, Song-Chun},
  title={{IconQA}: A New Benchmark for Abstract Diagram Understanding and Visual Language Reasoning},
  booktitle={Proceedings of the Neural Information Processing Systems Track on Datasets and Benchmarks},
  year={2021},
  url={https://iconqa.github.io/}
}

@inproceedings{AOKVQA,
  author={Schwenk, Dustin and Khandelwal, Apoorv and Clark, Christopher and Marino, Kenneth and Mottaghi, Roozbeh},
  title={{A-OKVQA}: A Benchmark for Visual Question Answering Using World Knowledge},
  booktitle={European Conference on Computer Vision (ECCV)},
  year={2022},
  url={https://arxiv.org/abs/2206.01718}
}

@inproceedings{ImageNetR,
  author={Hendrycks, Dan and Basart, Steven and Mu, Norman and Kadavath, Saurav and Wang, Frank and Dorundo, Evan and Desai, Rahul and Zhu, Tyler and Parajuli, Samyak and Guo, Mike and Song, Dawn and Steinhardt, Jacob and Gilmer, Justin},
  title={The Many Faces of Robustness: A Critical Analysis of Out-of-Distribution Generalization},
  booktitle={Proceedings of the IEEE/CVF International Conference on Computer Vision (ICCV)},
  pages={8340--8349},
  year={2021},
  url={https://arxiv.org/abs/2006.16241}
}

@inproceedings{Screen2Words,
  author={Wang, Bryan and Li, Gang and Zhou, Xin and Chen, Zhourong and Grossman, Tovi and Li, Yang},
  title={{Screen2Words}: Automatic Mobile {UI} Summarization with Multimodal Learning},
  booktitle={Proceedings of the 34th Annual ACM Symposium on User Interface Software and Technology},
  pages={498--510},
  year={2021},
  doi={10.1145/3472749.3474765},
  url={https://arxiv.org/abs/2108.03353}
}

@inproceedings{TabMWP,
  author={Lu, Pan and Qiu, Liang and Chang, Kai-Wei and Wu, Ying Nian and Zhu, Song-Chun and Rajpurohit, Tanmay and Clark, Peter and Kalyan, Ashwin},
  title={Dynamic Prompt Learning via Policy Gradient for Semi-structured Mathematical Reasoning},
  booktitle={The Eleventh International Conference on Learning Representations},
  year={2023},
  url={https://arxiv.org/abs/2209.14610}
}

@misc{Qwen35,
  author={{Qwen Team}},
  title={{Qwen3.5}: Towards Native Multimodal Agents},
  month={February},
  year={2026},
  url={https://qwen.ai/blog?id=qwen3.5}
}

@article{Qwen3VL,
  author={Shuai Bai and Yuxuan Cai and Ruizhe Chen and Keqin Chen and Xionghui Chen and Zesen Cheng and Lianghao Deng and Wei Ding and Chang Gao and Chunjiang Ge and Wenbin Ge and Zhifang Guo and Qidong Huang and Jie Huang and Fei Huang and Binyuan Hui and Shutong Jiang and Zhaohai Li and Mingsheng Li and Mei Li and Kaixin Li and Zicheng Lin and Junyang Lin and Xuejing Liu and Jiawei Liu and Chenglong Liu and Yang Liu and Dayiheng Liu and Shixuan Liu and Dunjie Lu and Ruilin Luo and Chenxu Lv and Rui Men and Lingchen Meng and Xuancheng Ren and Xingzhang Ren and Sibo Song and Yuchong Sun and Jun Tang and Jianhong Tu and Jianqiang Wan and Peng Wang and Pengfei Wang and Qiuyue Wang and Yuxuan Wang and Tianbao Xie and Yiheng Xu and Haiyang Xu and Jin Xu and Zhibo Yang and Mingkun Yang and Jianxin Yang and An Yang and Bowen Yu and Fei Zhang and Hang Zhang and Xi Zhang and Bo Zheng and Humen Zhong and Jingren Zhou and Fan Zhou and Jing Zhou and Yuanzhi Zhu and Ke Zhu},
  title={{Qwen3-VL} Technical Report},
  journal={arXiv preprint arXiv:2511.21631},
  year={2025},
  url={https://arxiv.org/abs/2511.21631}
}

@misc{Unsloth,
  author={Han, Daniel and Han, Michael and {Unsloth team}},
  title={{Unsloth}},
  year={2023},
  url={https://github.com/unslothai/unsloth}
}
\bibliographystyle{iclr2027_conference}

\end{document}